\documentclass[11pt,letterpaper]{article}
\usepackage{naaclhlt2016}
\usepackage{times}
\usepackage{latexsym}
\usepackage{url}
\usepackage{graphicx}
\usepackage{booktabs}

\naaclfinalcopy 
\def\naaclpaperid{***} 

\title{SemanticZ at SemEval-2016 Task 3:\\ Ranking Relevant Answers
in Community Question Answering\\ Using Semantic Similarity
Based on Fine-tuned Word Embeddings}

\author{Todor Mihaylov \\
Research Training Group AIPHES\\
Institute for Computational Linguistics\\
Heidelberg University\\
  {\tt mihaylov@cl.uni-heidelberg.de} \\\And
  Preslav Nakov \\
  Qatar Computing Research Institute, HBKU\\
  P.O. box 5825\\
  Doha, Qatar \\
  {\tt pnakov@qf.org.qa}\\}

\begin{document}
\maketitle
\begin{abstract}
We describe our system for finding good answers in a community forum, as defined in SemEval-2016, Task 3 on Community Question Answering. Our approach relies on several semantic similarity features based on fine-tuned word embeddings and topics similarities. In the main Subtask C, our primary submission was ranked third, with a MAP of 51.68 and accuracy of 69.94. In Subtask A, our primary submission was also third, with MAP of 77.58 and accuracy of 73.39.
\end{abstract}

\section{Introduction}
\label{sec:intro}
Posting questions that have already been asked in a community forum is annoying to users in the forum. The SemEval-2016 Task 3 on Community Question Answering\footnote{http://alt.qcri.org/semeval2016/task3/}\cite{nakov-EtAl:2016:SemEval} aims to solve this real-life problem. The main subtask (Subtask C) asks to find an answer that already exists in the forum and will be appropriate as a response to a newly-posted question. 
There is also a secondary, Subtask A, which focuses on Question-Comment Similarity and asks to rank the comments within a question-comment thread based on their relevance with respect to the thread's question.

Here, we examine the performance of using different word embeddings obtained with the Word2Vec tool \cite{mikolov-yih-zweig:2013:NAACL-HLT},
which we use to build vectors for the questions and the answers. We train classifiers using features derived from these embeddings to solve subtasks A and C.

Our contribution is in producing good word embeddings based on empirical evaluation of different configurations working in the Community Question Answering domain;  as they perform well, we make them freely available to the research community.\footnote{\url{https://github.com/tbmihailov/semeval2016-task3-cqa}}

\section{Related Work}
\label{sec:related}

This year's SemEval-2016 Task 3 is a follow up of  SemEval-2015 Task 3 on Answer Selection in Community Question Answering \cite{nakov-EtAl:2015:SemEval}.
The 2015 subtask A asked to determine whether an answer was relevant, potentially useful, or bad, while this year this is about ranking.

Here we focus on features that use semantic knowledge such as word embeddings, various features extracted from word embeddings, and topic models. Word embeddings and word embeddings similarities have been used by teams in the 2015 edition of the task \cite{belinkov-EtAl:2015:SemEval,VOLTRON2015,tran-EtAl:2015:SemEval,nicosia-EtAl:2015:SemEval}.  LDA topic have also been used \cite{tran-EtAl:2015:SemEval}.

Many other features have been tried for the task. For example, \newcite{tran-EtAl:2015:SemEval} used metadata about the question and the comment. User profile statistics such as number of \textit{Good}, \textit{Bad} and \textit{Potentially Useful} comments by a given user have been used to model user likelihood of posting different types of comment \cite{nicosia-EtAl:2015:SemEval}.
\newcite{vo-magnolini-popescu:2015:SemEval3} and \newcite{nicosia-EtAl:2015:SemEval} used syntactic tree similarities to compare questions to comments. The problem of selecting relevant answers has even been approached as a spam filtering task \cite{vo-magnolini-popescu:2015:SemEval3}.

\section{Data}
\label{sec:data}

In our experiments, we used annotated training, development and testing datasets,  as well as a large unannotated dataset, all provided by the SemEval-2016 Task 3 organizers. We further collected some additional unannotated in-domain data from some other sources, as explained below; finally, we used some models pretrained on out-of-domain data.

\textbf{Training, development, and testing data.} For Subtask A, there are  6,398 questions and 40,288 comments  from their question-answer threads, 
and for Subtask C, there are 317 original questions, 3,169 related questions, and 31,690 comments.
For both subtasks, the comments are annotated as \textit{Good}, \textit{PotentiallyUseful} and \textit{Bad}; for subtask A, the annotation is with respect to the question in whose thread the comment appeared, while for subtask C, it is with respect to a new question. For both subtasks, a successful ranking is one that ranks all \textit{Good} comments before all \textit{PotentiallyUseful} and \textit{Bad} ones (without distinguishing between the latter two).

\textbf{Unannotated data.} We performed experiments with Word2Vec embeddings trained on different unannotated data sources. We wanted to find the best performing embeddings and to use them in our system. In Table \ref{table:semantic-vectors-train-size}, we list the various data sources we used for training our Word2Vec models, and their vocabulary size.\\
\textbf{\textit{Qatar Living Forum}} is the original Qatar Living.\footnote{\url{www.qatarliving.com} is an online community for everyone living in or interested in the State of Qatar.} unannotated data containing 189,941 questions and 1,894,456 comments. It is limited to the forums section of the Qatar Living website.\\
\textbf{\textit{Qatar Living Forum + Ext}} includes the \textit{Qatar Living Forum} dataset, i.e., the forums, but also some other sections of Qatar Living: Jobs, Classifieds, Pages, Wiki and Events posts.\\
\textbf{\textit{Doha News}} is a dataset that we built by crawling about 7,000 news publications about the life in Doha, Qatar from the DohaNews website.\footnote{\url{dohanews.co} covers breaking news, politics, business, culture and more in and around Qatar.} 

We also used an out-of-domain, general model, readily-pretrained using Word2Vec on \textbf{\textit{Google News}},\footnote{\url{code.google.com/archive/p/word2vec/}} as provided by \newcite{mikolov-yih-zweig:2013:NAACL-HLT}.

\begin{table}[tbh]
\centering
\begin{tabular}{lcc}
\textbf{Features} & \textbf{Train size} & \textbf{Vocab}
\\\hline
Qatar Living Forum         & 61.84M & 104K\\
Qatar Living Forum+Ext & 90M & 126K\\
Google News             & 100B & 3M\\
Doha News               & 1.45M & 17K\\
\hline
\end{tabular}
\caption{\textbf{Data used for training word embedding vectors.} Shown are training source size (word tokens) and vocabulary size (word types).}
\label{table:semantic-vectors-train-size}
\end{table}

\section{Method}
\label{sec:method}

Below we focus our explanation on subtask A; for subtask C, we combine the predictions for subtask A with the Google's reciprocal rank for the related question (see below).

We approach subtask A as a classification problem. For each comment, we extract variety of features from both the question and the comment, and we train a classifier to label comments as Good or Bad with respect to the thread question. We rank the comments in each question according to the classifier's score of being classified as Good with respect to the question.

We first train several word embedding vector models and we fine-tune them using different configurations. For fine-tuning the parameters of the word embeddings training configuration, we setup a simple baseline system and we evaluate it on the official MAP score. We then use the best-performing embeddings in our further experiments. Our main features are semantic similarity based on word embeddings and topics, but we also use some metadata features. 

\subsection{Preprocessing}
Before extracting features, we preprocessed the input text using several steps. We first replaced URLs in text with TOKEN\_URL, numbers with TOKEN\_NUM, images with TOKEN\_IMG, and emoticons with TOKEN\_EMO. We then tokenized the text by matching only continuous alphabet characters including \_(underscore). Next, we lowercased the result. For the training, the development, and the test datasets, we removed the stopwords using the English stopwords lexicon from the NLTK toolkit \cite{LoperBird04}.

\subsection{Features}
We used several semantic vector similarity and metadata feature groups. For the similarity measures mentioned below, we used cosine similarity:

\begin{equation}
 1 - \frac{u.v}{\left\|u\right\|. \left\|v\right\|}
\label{formula:cosine-similarity}
\end{equation}

\textbf{Semantic Word Embeddings.}
We used semantic word embeddings obtained from Word2Vec models trained on different unannotated data sources including the QatarLiving and DohaNews. We also used a model pre-trained on Google News text. 
For each piece of text such as comment text, question body and question subject, we constructed the centroid vector from the vectors of all words in  that text (excluding stopwords).

\begin{equation}
centroid(w_{1..n}) = \frac{\sum\limits_{i=1}^n w_{i}}{n}
\label{formula:centroid-vector}
\end{equation}

We built centroid vectors (\ref{formula:centroid-vector}) from the question body and the comment text. We then examined different Word2Vec models in terms of training source and training configuration including word vector size, training window size, minimum word occurrence in the corpus, and number of skip-grams.

\textbf{Semantic Vector Similarities.} We used various similarity features calculated using the centroid word vectors on the question body, on the question subject and on the comment text, as well as on parts thereof:

\textbf{\textit{Question to Answer similarity.}} We assume that a relevant answer should have a centroid vector that is close to that for the question. We used the question body to comment text, and question subject to comment text vector similarities.

\textbf{\textit{Maximized similarity.}} We ranked each word in the answer text to the question body centroid vector according to their similarity and we took the average similarity of the top $N$ words. We took the top 1,2,3 and 5 words similarities as features. The assumption here is that if the average similarity for the top $N$ most similar words is high, then the answer might be relevant.

\textbf{\textit{Aligned similarity.}} For each word in the question body, we chose the most similar word from the comment text and we took the average of all best word pair similarities as suggested in \cite{tran-EtAl:2015:SemEval}. 

\textbf{\textit{Part of speech (POS) based word vector similarities.}} We performed part of speech tagging using the Stanford tagger \cite{Toutanova:2003:FPT:1073445.1073478}, and we took similarities between centroid vectors of words with a specific tag from the comment text and the centroid vector of the words with a specific tag from the question body text. The assumption is that some parts of speech between the question and the comment might be closer than other parts of speech.

\textbf{Word clusters (WC) similarity.} We clustered the word vectors from the Word2Vec vocabulary in 1,000 clusters (with 200 words per cluster on average) using K-Means clustering. We then calculated the cluster similarity between the question body word clusters and the answer text word clusters. For all experiments, we used clusters obtained from the Word2Vec model trained on QatarLiving forums with vector size of 100, window size 10, minimum words frequency of 5, and skip-gram 1.

\textbf{LDA topic similarity.} We performed topic clustering using Latent Dirichlet Allocation (LDA) as implemented in the \textit{gensim} toolkit \cite{rehurek_lrec} on Train1+Train2+Dev questions and comments. We built topic models with 100 topics. For each word in the question body and for the comment text, we built a bag-of-topics with corresponding distribution, and calculated similarity. The assumption here is that if the question and the comment share similar topics, they are more likely to be relevant to each other.

\textbf{Metadata.} In addition to the semantic features described above, we also used some common sense metadata features:

\textbf{\textit{Answer contains a question mark.}} If the comment has an question mark, it may be another question, which might indicate a bad answer.

\textbf{\textit{Answer length.}} The assumption here is that longer answers could bring more useful detail.

\textbf{\textit{Question length.}} If the question is longer, it may be more clear, which may help users give a more relevant answer.

\textbf{\textit{Question to comment length.}} If the question is long and the answer is short, it may be less relevant.

\textbf{\textit{The answer's author is the same as the corresponding question's author.}} If the answer is posted by the same user who posted the question and it is relevant, why has he/she asked the question in the first place?

\textbf{\textit{Answer rank in the thread.}} Earlier answers could be posted by users who visit the forum more often, and they may have read more similar questions and answers. Moreover, discussion in the forum tends to diverge from the question over time.

\textbf{\textit{Question category.}} We took the category of the question as a sparse binary feature vector (a feature with a value of 1 appears if question is in the category). The assumption here is that the question-comment relevance might depend on the category of the question.

\subsection{Classifier}
 For each Question+Comment pair, we extracted the features explained above from the \textit{Question} body and the subject text fields, and from the \textit{Comment} text; we also extracted the relevant metadata. We concatenated the extracted features in a bag of features vector, scaling them in the 0 to 1 range, and feeding them to a classifier. In our experiments, we used different feature configurations. We used L2-regularized logistic regression classifier as implemented in Liblinear \cite{liblinear}. For most of our experiments, we tuned the classifier with different values of the C (cost) parameter, and we took the one that yielded the best accuracy on 5-fold cross-validation on the training set. We used binary classification \textit{Good} vs. \textit{Bad} (including both \textit{Bad} and \textit{Potentially Useful} original labels). The output of the evaluation for each test example was a label, either \textit{Good} or \textit{Bad}, and the probability of being \textit{Good} in the 0 to 1 range. We then used this output probability as a relevance rank for each \textit{Comment} in the \textit{Question} thread.

\section{Experiments and Evaluation}
\label{sec:experiments}

As explained above, we rely mainly on semantic features extracted from Word2Vec word embeddings. Thus, we ran several experiments looking for the best embeddings for the task.

\begin{table}[h!]
\centering
\begin{tabular}{lcc}
 & \multicolumn{2}{c}{ \bf Dev2016} \\
\textbf{Dataset} & \textbf{MAP} & \textbf{Accuracy}
\\\hline
Qatar Living Forum         & \bf 0.6311 & 0.7078\\
Qatar Living Forum+Ext & 0.6269 & \bf 0.7131\\
Google News             & 0.6113 & 0.6996\\
Doha News               & 0.5769 & 0.6844\\
\hline
\end{tabular}
\caption{\textbf{Semantic vectors trained on different unannotated datasets as the only features for subtask A:} training on train2016-part1, testing on dev2016.}
\label{table:experiments-semantic-vectors}
\end{table}

\begin{table}[h!]
\centering
\begin{tabular}{ccc}
  & \multicolumn{2}{c}{ \bf Test2016} \\
\textbf{Vector size} & \textbf{MAP} & \textbf{Accuracy}
\\\hline
800 & \bf 78.45 & 74.22\\
700 & 78.12 & 73.98\\
600 & 77.31 & 73.15\\
500 & 77.61 & 73.30\\
400 & 78.36 & 74.19\\
300 & 77.25 & 74.50\\
200 & 77.90 & 73.88\\
100 & 77.08 & \bf 74.53\\
50 & 77.22 & 73.85\\
20 & 75.44 & 72.42\\\hline
Baseline & 59.53 & -\\
\end{tabular}
\caption{\textbf{Semantic vectors of different vector sizes, trained on Qatar Living Forum+Ext as features for subtask A (together with all other features):} training on train2016-part1, testing on test2016.}
\label{table:experiments-vectors-train-params-size}
\end{table}

Table \ref{table:experiments-semantic-vectors} shows experiments with Word2Vec models trained on the unannotated datasets described above. The Google News Word2Vec model comes pretrained with vector size of 300, window 10, minimum word frequency of 10 and skip-gram 1. We started with training our three Word2Vec models using the same parameters. 

Table \ref{table:experiments-semantic-vectors} shows results using raw word vectors as features, together with an extra feature for question body to comment cosine similarity. We can see that training on \textit{Qatar Living Forum} data performs best followed by using \textit{Qatar Living Forum+Ext}, \textit{Google News}, and \textit{Doha News}. This is not surprising as the first two datasets are in-domain, while the latter two cover more topics (as they are news) and more formal language. Overall, Doha News contains topics that largely overlap with the topics discussed in the Qatar Living forum; yet, it uses more formal language and contains very little conversational word types (mostly in quotations and interviews); moreover, being smaller in size, it covers much less vocabulary. Based on these preliminary experiments on Dev2016, we concluded that the domain-specific word vectors trained on \textit{Qatar Living Forum} were the best for this task, and we used them further in our experiments. 

After we have selected the best dataset for training our semantic vectors, we continued with various experiments to select the best training parameters for Word2Vec.
Below we present the results of these experiments on Test2016, but we experimented with Dev2016 when developing our system.

In Table \ref{table:experiments-vectors-train-params-size}, we present experiments with different vector sizes. We trained our classifier with all features mentioned above, extracted for the corresponding word vector model. We can see that word vectors of size 800 perform best followed by sizes 400 and 700. However, we should note that using word vectors of size 800 generates more than 1,650 features (800+800+other features), which slows down training and evaluation. 
Moreover, in our experiments, we noticed that using large word vectors blurs the impact of the other, non-vector features.

Thus, next we tried to achieve the MAP for the 800-size vector by using better parameters for smaller vector sizes. Table \ref{table:experiments-vectors-grid-search} shows the results, where we used vectors of size 100 and 200. 
We can see that the configuration with word vector size 200, window size 5, minimum word frequency 1 and skip-gram 3 performed best improving the 200 vectors MAP by 0.31 (compared to Table \ref{table:experiments-vectors-train-params-size}). 
However, the experiments with word vector size 100 improved its MAP score by 0.85, which suggests that there might be potential for improvement when using vectors of smaller size. We also tried to use Doc2Vec \cite{DBLP:journals/corr/LeM14} instead of Word2Vec, but this led to noticeably lower performance.

\begin{table}[tbh]
\centering
\begin{tabular}{cccccc}
  & & & & \multicolumn{2}{c}{ \bf Test2016} \\
\bf Size & \bf Window & \bf Freq & \bf Skip & \textbf{MAP} & \textbf{Acc}
\\\hline
 200 &   5 &   1 &   3 & \bf 78.21 & 74.25\\
 200 &   5 &   5 &   1 & 78.19 & 73.49\\
 200 &   5 &   5 &   3 & 78.13 & 74.01\\
 200 &   5 &   1 &   1 & 78.01 & \bf 74.53\\
 100 &   5 &   1 &   1 & 77.93 & 74.19\\
 200 &   10 &   5 &   1 & 77.90 & 73.88\\
 100 &   5 &   1 &   3 & 77.81 & 73.94\\
 100 &   10 &   1 &   1 & 77.72 & 74.43\\
 200 &   10 &   1 &   1 & 77.58 & 74.25\\
 100 &   5 &   5 &   1 & 77.53 & 74.07\\
 200 &   10 &   1 &   3 & 77.43 & 73.73\\
 100 &   10 &   10 &   1 & 77.18 & 73.79\\
 100 &   10 &   5 &   1 & 77.08 & \bf 74.53\\
\hline
\end{tabular}
\caption{\textbf{Exploring Word2Vec training parameters on Qatar Living Forum+Ext:} word vector size (Size), context window (Window), minimum word frequency (Freq), and skip-grams (Skip).
Vectors used as features for subtask A (together with all other features):
training on train2016-part1, testing on test2016.}
\label{table:experiments-vectors-grid-search}
\end{table}

\begin{table}[h!]
\centering
\begin{tabular}{lcccc}
 \bf Train2016-part1 as training & \multicolumn{2}{c}{ \bf Test2016} \\
\textbf{Features} & \textbf{MAP} & \textbf{Acc} \\
\hline
All $-$ Quest. to Comment sim                 & \bf 78.52 & 74.31 \\
All $-$ Maximized similarity       & 78.38 & 74.59 \\
All $-$ Word Clusters similarity   & 78.29 & 74.25 \\
All $-$ WC sim \& Meta cat        & 78.22 & 74.04 \\
All $-$ Meta categories            & 78.21 & 74.25 \\
All                                & 78.21 & 74.25 \\
All $-$ Meta cat \& LDA sim           & 78.18 & 73.88 \\
All $-$ Ext POS sim \& WC sim     & 78.10 & 74.28 \\
All $-$ Aligned similarity         & 77.97 & 74.16 \\
All $-$ Cat \& WC \& LDA sim & 77.95 & 74.19\\
All $-$ WC \& LDA sims         & 77.92 & 74.25 \\
All $-$ Ext POS sim                & 77.92 & 74.43 \\
All $-$ LDA sim                    & 77.85 & 74.37 \\
All $-$ POS sim                    & 77.77 & \bf  74.80 \\
All $-$ Metadata full              & 74.50 & 70.31 \\
All $-$ Word Vectors               & 74.35 & 70.80 \\\hline
Primary                            & 77.58 & 73.39 \\
Contrastive 1                      & 77.16 & 73.88 \\
Contrastive 2                      & 75.41 & 72.26 \\\hline
Baseline (IR)  & 59.53 & -- \\\hline
\end{tabular}
\caption{\textbf{Subtask A. Using all features without some feature groups.}
Word2Vec is trained with word vector size 200, context window 5, minimum word frequency 1, and skip-grams 3.}
\label{table:subtask-a-allfeatures}
\end{table}

\begin{table}[tbh]
\centering
\begin{tabular}{lcccc}
\bf Train2016-part1 as training & \multicolumn{2}{c}{ \bf Test-2016} \\
\textbf{Features} & \textbf{MAP} & \textbf{Acc}\\
\hline
All $-$ Q to C sim              & \bf 53.39 & 69.87 \\
All $-$ Meta categories         & 53.06 & 69.81 \\
All $-$ WC sim \& Meta cat      & 52.91 & 69.54 \\
All $-$ WC sim \& LDA sim       & 52.84 & 70.06 \\
All $-$ Meta cat \& LDA sim         & 52.83 & 69.87 \\
All $-$ Ext POS sim \& WC sim   & 52.82 & 70.21 \\
All                             & 52.78 & 69.43 \\
All $-$ Aligned similarity      & 52.76 & 70.10 \\
All $-$ Word Clusters similarity & 52.58 & 69.63 \\
All $-$ Maximized similarity    & 52.47 & 69.27 \\
All $-$ Cat \& WC \& LDA sim & 52.44 & 69.51 \\
All $-$ Exr POS sim             & 52.23 & 69.91 \\
All $-$ LDA sim                 & 52.08 & 69.97 \\
All $-$ POS sim                 & 51.57 & 69.96 \\
All $-$ Word Vectors            & 49.57 & 70.13 \\
All $-$ Metadata full           & 46.03 & \bf 71.06 \\\hline
Primary & 51.68 & 69.94         \\
Contrastive 1                   & 51.46 & 69.69 \\
Contrastive 2                   & 48.76 & 69.71 \\\hline
Baseline (IR) & 28.88 & --\\\hline
\end{tabular}
\caption{\textbf{Subtask C. Using all features without some feature groups.}
Word2Vec is trained with word vector size 100, context window 5, minimum word frequency 1, and skip-grams 1.}
\label{table:subtask-c-allfeatures}
\end{table}

We further experimented with Word2Vec models trained with different configurations and different feature groups. 
Tables \ref{table:subtask-a-allfeatures} and \ref{table:subtask-c-allfeatures} show the results for ablation experiments using the best-performing configuration for Subtask A and C, respectively.

For Subtask A we achieved the best score with semantic vectors of size 200, trained with window size 5, minimum word frequency 1 and skip-grams 3. The best score we achived (MAP 78.52) is slightly hbetter than the best score from Table~\ref{table:experiments-vectors-train-params-size} (MAP 78.45), which means that it may be a good idea to use smaller word vectors in combination with other features. We can see that the features that contribute most (the bottom features are better) are the raw word centroid vectors and metadata features, followed by various similarities such as LDA topic similarity and POS-tagged-word similarity.

For Subtask C, we achieved the best score with vectors of size 100, trained with window size 5, minimum word frequency 1, and skip-grams 1. The features that contributed most were mostly the same as for Subtask A. One difference is the maximized similarity features group, which now yields worse results when excluded, which indicates its importance.

Our \textit{Primary}, \textit{Contrastive 1} and \textit{Contrastive 2} submissions were built with the same feature set: \textit{All features - POS similarity \& Meta Category}, but were trained with fixed C=0.55 on different datasets: \textit{Primary} was trained on Train2016-part1, \textit{Contrastive 1} was trained on Train2016-part1 + Train2016-part2, and \textit{Contrastive 2} was trained on Train2016-part2.

\section{Conclusion and Future Work}
\label{sec:future}

We have described our system for SemEval-2016, Task 3 on Community Question Answering. Our approach relied on several semantic similarity features based on fine-tuned word embeddings and topics similarities. 

In the main Subtask C, our primary submission was ranked third, with a MAP of 51.68 and accuracy of 69.94. In Subtask A, our primary submission was also third, with MAP of 77.58 and accuracy of 73.39.
After the submission deadline, we improved our MAP score to 78.52 for Subtask A, and to 53.39 for Subtask C, which would rank our system second.

In future work, we plan to use our best performing word embeddings models and features in a deep learning architecture, e.g., as in the MTE-NN system \cite{ACL2016:MTE-NN-cQA,SemEval2016:task3:MTE-NN}, which borrowed an entire neural network framework and achitecture from previous work on machine translation evaluation \cite{guzman-EtAl:2015:ACL-IJCNLP}. 
We also want to incorporate several rich knowledge sources, e.g., as in the SUper Team system \cite{SemEval2016:task3:SUper},
including 
troll user features as inspired by \cite{mihaylov-georgiev-nakov:2015:CoNLL,mihaylov-EtAl:2015:RANLP2015,ACL2016:trolls},
and PMI-based goodness polarity lexicons as in the PMI-cool system \cite{SemEval2016:task3:PMI-cool},
as well as sentiment polarity features \cite{nicosia-EtAl:2015:SemEval}.

We further plan to use information from entire threads 
to make better predictions, as using thread-level information for answer classification has already been shown useful for SemEval-2015 Task 3, subtask A, e.g., by using features modeling the thread structure and dialogue~\cite{nicosia-EtAl:2015:SemEval,barroncedeno-EtAl:2015:ACL-IJCNLP}, or by applying thread-level inference using the predictions of local classifiers~\cite{joty:2015:EMNLP,Joty:2016:NAACL}. 
How to use such models efficiently in the ranking setup of 2016 is an interesting research question.

Finally, we would like to address subtask C in a more solid way, making good use of the data, the gold annotations, the features, the models, and the predictions for subtasks A and B.

\textbf{Acknowledgments.} This work is partly supported by the German Research Foundation as part of the Research Training Group ``Adaptive Preparation of Information from Heterogeneous Sources'' (AIPHES) under grant
No. GRK 1994/1. 
It is also part of the Interactive sYstems for Answer Search (Iyas) project, which is developed by the Arabic Language Technologies (ALT) group at the Qatar Computing Research Institute, HBKU, part of Qatar Foundation in collaboration with MIT-CSAIL.

\bibliography{bib}
\bibliographystyle{naaclhlt2016}

\end{document}